\documentclass[conference]{IEEEtran}
\IEEEoverridecommandlockouts
\usepackage{cite}
\usepackage{url}
\usepackage{amsmath,amssymb,amsfonts}
\usepackage{algorithmic}
\usepackage{graphicx}
\usepackage{subcaption}
\usepackage{float}
\usepackage{textcomp}
\usepackage{xcolor}
\usepackage{comment}
\usepackage{parskip}
\usepackage{adjustbox}

\def\BibTeX{{\rm B\kern-.05em{\sc i\kern-.025em b}\kern-.08em
    T\kern-.1667em\lower.7ex\hbox{E}\kern-.125emX}}

\begin{document}
\title{A Beacon Based Solution for Autonomous UUVs GNSS-Denied Stealthy Navigation\\

\thanks{Research funding for this project was provided by the French National Research Agency via the project TSIA NAIAD (ANR-23-IAS2-0006-01).}
}

\author{\IEEEauthorblockN{Alexandre Albore}
\IEEEauthorblockA{\textit{DTIS ONERA, Universit\'e de Toulouse} \\
Toulouse, France \\
alexandre.albore@onera.fr}
\and
\IEEEauthorblockN{Humbert Fiorino}
\IEEEauthorblockA{\textit{LIG, Universit\'e de Grenoble Alpes}\\
Grenoble, France \\
humbert.fiorino@univ-grenoble-alpes.fr}
\and
\IEEEauthorblockN{Damien Pellier}
\IEEEauthorblockA{\textit{LIG, Universit\'e de Grenoble Alpes}\\
Grenoble, France \\
damien.pellier@imag.fr}
}

\maketitle

\begin{abstract}
Autonomous Unmanned Underwater Vehicles (UUVs) enable military and civilian covert operations in coastal areas without relying on support vessels or Global Navigation Satellite Systems (GNSS). Such operations are critical when surface access is not possible, and stealthy navigation is required in restricted environments -- such as protected zones, or dangerous area under access ban. GNSS-denied navigation is then essential to maintaining concealment, as surfacing could expose UUVs to detection.

To ensure a precise fleet positioning, a constellation of beacons deployed by aerial or surface drones establish a synthetic landmark network, that will guide the fleet of UUVs along an optimized path from the continental shelf to the goal on the shore. These beacons, either submerged or floating, emit acoustic signals for UUV localisation and navigation. A hierarchical planner generates an adaptive route for the drones, executing primitive actions while continuously monitoring and replanning as needed to maintain trajectory accuracy.
\end{abstract}


\section{Introduction}

Military and civil discreet operations can be carried out by deploying a fleet of autonomous underwater drones in coastal areas, without immediate access to a support vessel, which remains in the open sea, and without access to Global Navigation Satellite Systems (GNSS).
This application is typical of areas where mapping or monitoring of natural phenomena or pollution is required, without access for the carrier because of submerged bottom, a protected natural area, or even a dangerous area under access ban, etc.
GNSS-denied underwater navigation is a necessity that arises in both civilian and military contexts, when surfacing will compromise the concealment provided by the underwater environment, because the Unmanned Underwater Vehicles (UUVs) would be detectable by visual, radar, and other surveillance systems.
This necessity may arise in civilian contexts, such as environmental monitoring, marine biology, delivering essential supplies through humanitarian corridors, or offshore industry operations; precise and discreet navigation ensures that missions are conducted without disturbing the natural environment or sensitive sites. For military applications, navigating stealthily and without GNSS is central for mapping and intelligence gathering, and protecting underwater infrastructures.

The development and integration of advanced navigation technologies are critical to the success of these missions, ensuring that UUVs can operate effectively underwater.

Despite the centrality of this challenge and the interest of industrial research, GNSS-denied navigation for marine and underwater vehicles remains rare in publicly available research, particularly when compared to the mature GNSS-denied alternative localization systems seen in Unmanned Aerial Vehicles (UAVs) and terrestrial robots.


One of the main reasons GNSS cannot be used underwater to locate a target is that GPS is unavailable due to the rapid attenuation of high frequency electromagnetic waves underwater \cite{masmitja2019range}.

Economic, industrial, and environmental strategies that control submarine resources face several difficulties. Given that a fleet of unmanned vehicles is a relevant solution for performing inspection, cartography, and interventions, its reliable and robust deployment requires solving several issues. On one side, the environment is not always fully known, and is strongly dynamic, which affects the capacity of the UUVs in navigating. On the other side, GNSS is not available when drones are immersed, which affects the correct localization of the fleet and thus its navigation robustness.

To address this problem, the most common approaches use acoustic localization systems, sometimes performing multi-source data and sensor fusion.

Paull et al. \cite{paull2014auv} summarize common localization methods using acoustic time-of-flight measurements.
Underwater navigation employs various acoustic methods for localization and tracking. Ultra-short baseline and short baseline systems operate from ships, using time-of-flight and phase differences to calculate range and azimuth, but face limitations in range and accuracy. Long baseline (LBL) systems use widely spaced fixed beacons for precise acoustic triangulation, though deployment and calibration remain a complex task. Single fixed beacon systems offer a cost-effective alternative, but lack unique position solutions. Both LBL and single beacon systems rely on two-way travel time of acoustic signals. Additionally, mobile underwater vehicles equipped with sensors can serve as an alternative for target tracking.


Through-the-sensor (TTS) sub-bottom imaging~\cite{sabra2025TTS} leverages the autonomous UUV's inherent self-noise as an acoustic source for mapping beneath the ocean floor. By utilizing the low-frequency noise produced by the vehicle's propulsion system, this technique offers a stealthier and more energy-efficient alternative to conventional sonar methods, enabling continuous and covert subsurface imaging without the need for active sonar pings or external emitters.

Most approaches for autonomous UUVs aim at detecting a moving underwater vehicle.
Masmitja et al. \cite{Masmitja18} studied the optimal path for a mobile robot localizing a static target with the range-only and single-beacon (ROSB) method,  based on an autonomous vehicle that localizes and tracks different underwater targets using slant range measurements carried out with acoustic modems. This approach is an alternative to more traditional approaches, such as LBL. The simulation and in-water results for the ROSB method with a wave-glider determined that a circular USV/target path was optimal.


We tackle this navigation problem using a different approach:
to meet the need for accurate location, it is envisaged to use external devices -- in this case, positioning beacons dropped by an aerial or surface drone deployed from a drone-carrying vessel sailing off the coast. The purpose of this constellation of synthetic landmarks is to mark out an optimal navigation plan, considering the constraints of the environment and the means deployed.

Once the path is marked by the beacons, the fleet will direct itself following the locator beacons, from the initial position at the limit of the continental shelf to the goal on a beach. We assume that the locator beacons were deployed by a UAV or a surface vehicle. They were dropped into the water and can thus sink to the seafloor, float midwater, or float like a buoy on the surface, depending on their characteristics. The beacons emit an acoustic pulse intended to guide UUVs equipped with a suitable receiver.
Finally, the UUVs navigation plan is generated by a hierarchical planner, that provides a sequence of primitive actions to follow the marked trajectory. The hierarchical planning model used will provide the best adapted course of actions, given the state of the localization of the fleet of UUVs.

In the paper, we first describe the details of the use case in Section~\ref{sec:UC}, then in Section~\ref{sec:archi} we describe the overall architecture of the framework and detail the hierarchical planning approach to synthesize the navigation plan; Section~\ref{sec:Algo} describes the beacons deployment algorithm developed, and its planning counterpart, where an automated planner produces a hierarchical plan for the UUVs to follow the marked path.
In Section \ref{sec:Implem}, we discuss the implementation of the algorithms as a QGIS plugin, and its execution on a simple fleet navigation example. In Section \ref{sec:conclu}, we conclude and comment on future developments of this approach.

\section{Use case}
\label{sec:UC}

This use case stands on the context of the control of the underwater domain. It adopts robotic solutions to perform surveillance and intervention operations.
The aim is to increase the decision-making autonomy of a fleet of unmanned vehicles and to enhance the robustness of underwater operations, even without an underwater satellite positioning system and without immediate access for a carrier vessel.

The main problem in carrying out navigation operations is the ability for each drone to reach the area of interest and to be able to position itself precisely given the inherent limits of submarine communication due to the very nature of the underwater environment that will affect the beacons signals and the inter-UUVs communications. As written above, many environmental hindrances can affect the success of the mission, namely: the presence of a potentially variable current in the area of operation, the limited performance of navigation equipment, the lack of bottom detection at the beginning of the survey, making it impossible to use a Doppler Velocity Logger (DVL), the impossibility for underwater drones to surface to take a GNSS fix, and the extremely limited underwater communications.
In order to circumvent these issues, this project aims at integrating different methodologies from the field of artificial intelligence with technological solutions implemented by a fleet of autonomous robots navigating in an unknown environment.

The localization is provided by a constellation of beacons able to transmit an acoustic signal up to 2000~m at a very low data rate (using HF acoustic waves). Each UUV in our scenario is equipped for navigation with an acoustic modem, a DVL, an AIS (Automatic Identification system), a camera, GPS, 4G, and an Inertial Navigation System (INS). 

The mission operation carried out by the fleet comprises reaching a given position using the beacons for localization support.
A UUV can broadcast messages to the fleet and sense a beacon's acoustic signal if it is within its range.
The navigation accuracy depends on the availability and usage of the signals from the beacons, as well as the influence of environmental factors.

For the first implementation of the navigation framework, and given the complexity of the problem,  we used some strong assumptions in the model and the uncertainty propagation. In particular, we use a simplified ocean environment where signal attenuation due to the different ocean layers is ignored. The absorption of sound in water increases with frequency, which limits the distance and bandwidth of submarine communication, and sound waves can also be refracted and scattered by different ocean layers, temperature gradients, and water currents.
%
The planning and monitoring are not centralized, but performed in each UUV, considered as an independent agent.

\section{Software Architecture}
\label{sec:archi}
\begin{figure}[th]
\centering
\includegraphics[width=.5\textwidth]{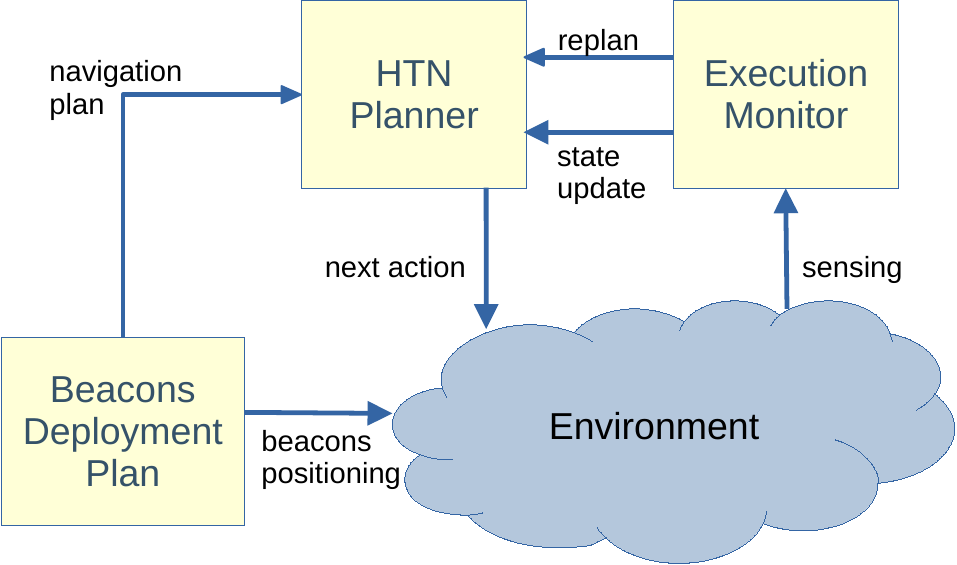}
\caption{The reactive planning framework. The plan is executed in a closed-loop fashion, where each action is applied in the environment and a monitor gathers the sensors outcome and triggers a replanning episode if needed.}\label{fig:arch}
\end{figure}

It is worth describing the overall software architecture of the planning and execution framework for the fleet of UUVs (Figure~\ref{fig:arch}), including the beacons deployment plan, before digging into the single elements of the tool suite.

The problem of placing beacons to mark the navigation path from the support vessel to the goal along the coastline is implemented as a QGIS plugin~\cite{QGISsoftware}, while the navigation plan representation is implemented as another QGIS plugin named TraceQGIS, described later in Section~\ref{sec:Implem}.
The beacon positioning mission is carried out before deploying the fleet, by aerial or surface drones.

The synthesis of the UUV trajectory is done by an automated hierarchical planner, on the basis of the position of the beacons.
A planner is a software that, given a model of the environment, solves an automated planning problem and provides a set of actions to accomplish a task, which in our case is the UUV fleet mission.

More in general, Automated Planning (also known as AI Planning) is the branch of semantic Artificial Intelligence that focuses on the development of strategies or sequences of actions to achieve specific goals~\cite{ghallab2016automated}.
Within the class of intelligent autonomous systems, AI planning is the paradigm that uses a model (of the environment, of the autonomous agents, \ldots) to perform this decision-making task that is action selection.
A planner, i.e. an automated solver for classes of mathematical models represented in compact form, provides as output a set of actions to drive the initial state of the system into a final, desired, goal state~\cite{geffner13}.
This approach is typically used by intelligent agents, autonomous robots, and unmanned vehicles.

The automatic synthesis of mission plans for the UUVs described here are modelled according to the Hierarchical Task Network (HTN) planning paradigm~\cite{erol1994umcp,GEORGIEVSKI2015124}, in order to break down the mission into elementary tasks, which allows a rational use of resources, and to generate alternative decompositions for the same mission task.

Several high level actions are possible: navigating on a course, sensing a beacon, broadcasting a message.

The navigation trajectory corresponding to the synthetic landmarks is the input of a hierarchical planner.

HTN planning models the AI planning problem as a set of complex abstract tasks to solve, each one of them broken down into (possibly different) specific subtasks, down to the primitive actions.
HTN planning has initially been developed with planners such as NOAH~\cite{sacerdoti1975}, SIPE~\cite{wilkins1990}, and UMCP~\cite{erol1994umcp}. For totally ordered HTNs, it is possible to plan tasks in the  order they will be executed, and thus to know the current  state of the environment at each step of the planning process.

The HTN planner synthesizes the navigation plan that has the main objective of guiding the UUVs from beacon to beacon, until reaching the mission goal.
Indeed, the navigation planning problem is subject to multiple sources of uncertainty: beacons can be displaced due to ocean currents, intermittent signal transmission, and UUV trajectory perturbations caused by subsurface dynamics. Thus, simply adhering to directions traced by the beacons cannot guarantee a reliable path following.
The plan is then executed in a closed loop fashion, where an execution monitor gathers navigation data to elaborate the current state of the UUV,  providing a continuous assessment of the state of each element of the fleet. Close-loop plan execution is a common technique for environments where unexpected events can occur \cite{FF-replan,Karpas2020-PlanningforRobotics}.
If the current state does not match the  state expected on the basis of the known information about the environment and the initial assumptions, e.g. a beacon signal is not detected when expected, then a replanning episode is triggered for all the elements of the fleet that are in communication range, starting from the current state and adapted to react to changes in the environment and hazards of the mission.
 Note that ideally, replanning should occur for the whole fleet, but in a distributed environment, where communication is not guaranteed, such an optimization is not possible.

\section{Beacons deployment and Hierarchical Planning Navigation}
\label{sec:Algo}

The aim of deploying beacons in a given volume of water is to place them so to maximize the information gathered by the sensors. The UUV sensors provide a binary response, indicating whether the beacon signal is detected or not.

Considering a single UUV, the imperfect or unknown information on its current position -- due to currents, signal reflection and scattering, etc. -- implies that its position is known within a certain degree of uncertainty.
In a first approximation, in order to maximize the probability of gathering useful information during navigation, we minimize the average absolute deviation between the volume of water where the beacon signal can be received, and the volume of water where it cannot:

\begin{figure*}[h!t]
    \centering
    \includegraphics[width=\linewidth]{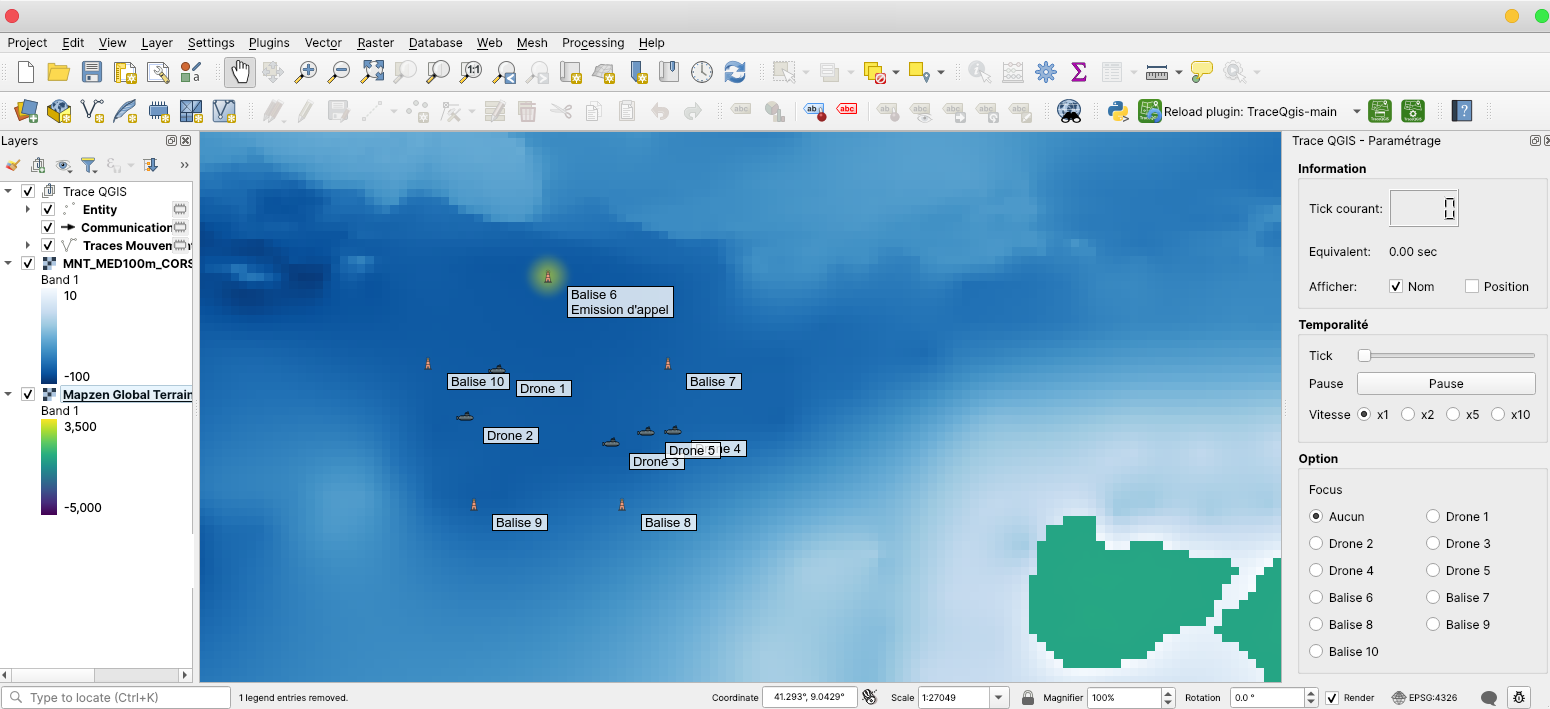}
    \caption{QGIS plugin interface featuring the beacons, and the UUVs off the coast of Capo Testa, Sardinia. On the right, the planning interface, with the time cursor, and the option to focus on the entities present on the map. On the left, the layers for the mission, and the bathymetric data.}
    \label{fig:interface}
\end{figure*}
\begin{equation}
\min \frac{1}{N} \sum_{n \leq N} \left| V_n - \frac{V_{tot}}{N} \right| \label{eq}
\end{equation}
 where $V_n$ is the volume of water around each beacon $n$, $N$ is the total number of beacons, and $V_{tot}$
 the total volume of water under the polygon delimiting the deployment area.

From this consideration, we implemented the Constrained Lloyd's algorithm with explicit volume targets equal to $\frac{V_{tot}}{N}$, where $V_{tot}$ corresponds to the volume of water below a polygon initially drawn on a bathymetric map (this quantity can easily be calculated within QGIS) and that corresponds to the area where the mission will be executed.
The algorithm iteratively distributes points within the given volume of water, splitting it in $N$ zones of equal volume. During each iteration of the algorithm, Lloyd iteration builds a Voronoi map in which each point is contained within a distinct Voronoi cell, then centres each point within its cell~\cite{lloyd82}. The centroid of each cell corresponds to the beacon's deployment position.

 The result is a set of beacon deployment positions in the mission area, defined by the polygon that encompasses the volume $V_{tot}$.


\subsection*{HTN Automated planning}
\label{sec:Planning}

For the execution of the UUVs mission, we consider synthesizing plans under the Hierarchical Task Network (HTN) planning paradigm~\cite{sacerdoti1975,erol1994umcp,GEORGIEVSKI2015124} in a complex marine environment. This approach allow to use different degrees of granularity to describe the planning problem.

 A key concept in HTN planning is the {\it task}.
 In HTN planning, we distinguish two kinds of tasks: the  primitive tasks (also called actions), and the abstract tasks (or compound tasks).
 Primitive tasks are carried out by actions in the sense of classical planning \cite{fikes71}, while abstract tasks can be refined by applying methods that define the decomposition of the task into subtasks. The purpose of abstract tasks is not to induce a state transition, while actions do change the state of the world. Instead, abstract tasks refer to a predefined mapping to one or more tasks that can refine the abstract task. For instance, in the task of serving a dinner, {\it deliver-dinner(?food-style, ?place)} is the compound task consisting in performing first the task of serving the starters, then the main course, etc. In that sense,  {\it deliver-dinner(?food-style, ?place)} can be refined in: $\langle${\it serve-starters(?food-style, ?place)},  {\it serve-main-course(?food-style, ?place)}, etc.$\rangle$



Let us first define the planning domain and problem for  HTN Planning.
 %
   A \emph{planning domain} $\cal{D}$ is a tuple $({\cal{L}}, {\cal{T}}, {\cal{J}}, \alpha, {\cal{A}}, {\cal{M}})$, where
   $\cal{L}$ is the first-order logic language,
   $\cal{T}$ is the set of tasks,
   $\cal{J}$ is the set of task identifiers\footnote{Task identifiers are arbitrary symbols, which serve as place holders for the actual tasks they represent. Identifiers are needed because tasks can occur multiple times within the same task network, as we will see below.},
   $\alpha: {\cal{J}} \rightarrow {\cal{T}}$ is the function that maps task identifiers to tasks,
   ${\cal{A}}$ is a set of actions,
   $\cal{M}$ is the set of methods. 
 The domain implicitly defines the set of all states $S$, defined over all subsets of all ground predicates in $\cal{L}$.

Tasks are connected together in HTN planning by task networks (TNs).
A TN is given by the pair $\bigl\{ \{ n_i \}_i, \Phi \bigr\}$, i.e. an ordered set of task identifiers $n_i  \in \cal{J}$ and the constraints $\Phi$ on those tasks.
This mapping between tasks is achieved by a set of decomposition methods; a method is then defined by the pair $\{ t, w \}$, with $t \in \cal{T}$ the task it decomposes, and $w$ the task network defining the decomposition into sub-tasks.

\begin{figure*}[!t]
\subfloat[UUV 1 navigates towards North-East.]{%
\includegraphics[width=.48\linewidth]{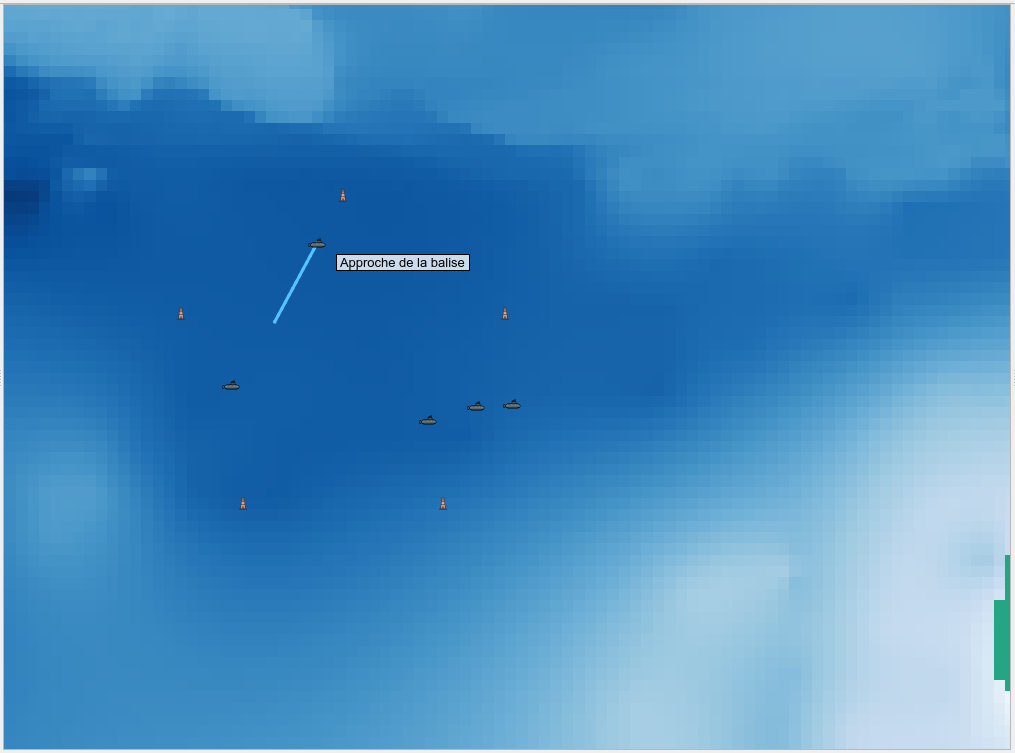}%
\label{subfig:a}%
}\hfill
\subfloat[UUV 1 receives beacon 6 signal.]{%
\includegraphics[width=.48\linewidth]{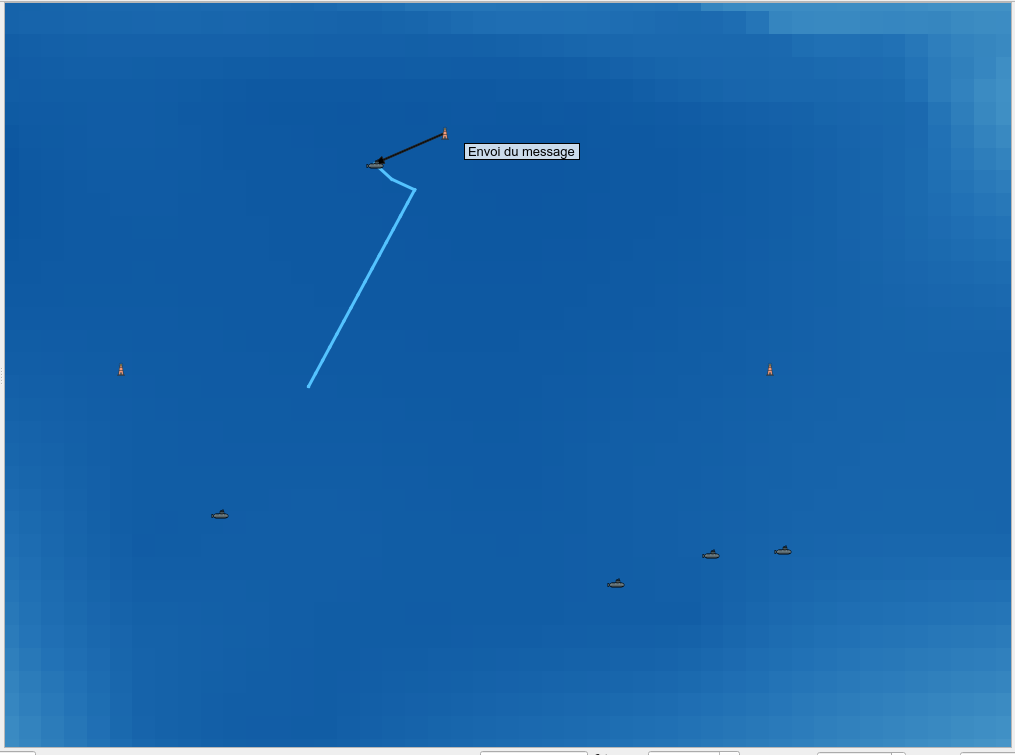}%
\label{subfig:b}%
}\\
\subfloat[UUV 1 broadcasts to the fleet.]{%
\includegraphics[width=.48\linewidth]{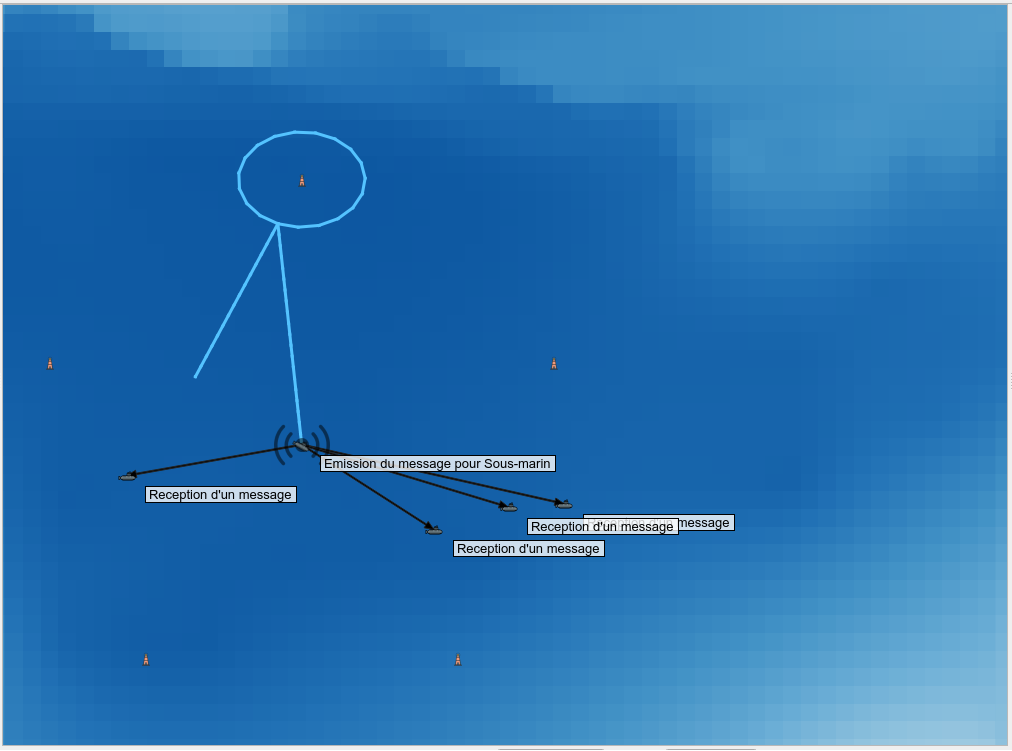}%
\label{subfig:c}%
}\hfill
\subfloat[UUV 1 navigates towards beacon 8.]{%
\includegraphics[width=.48\linewidth]{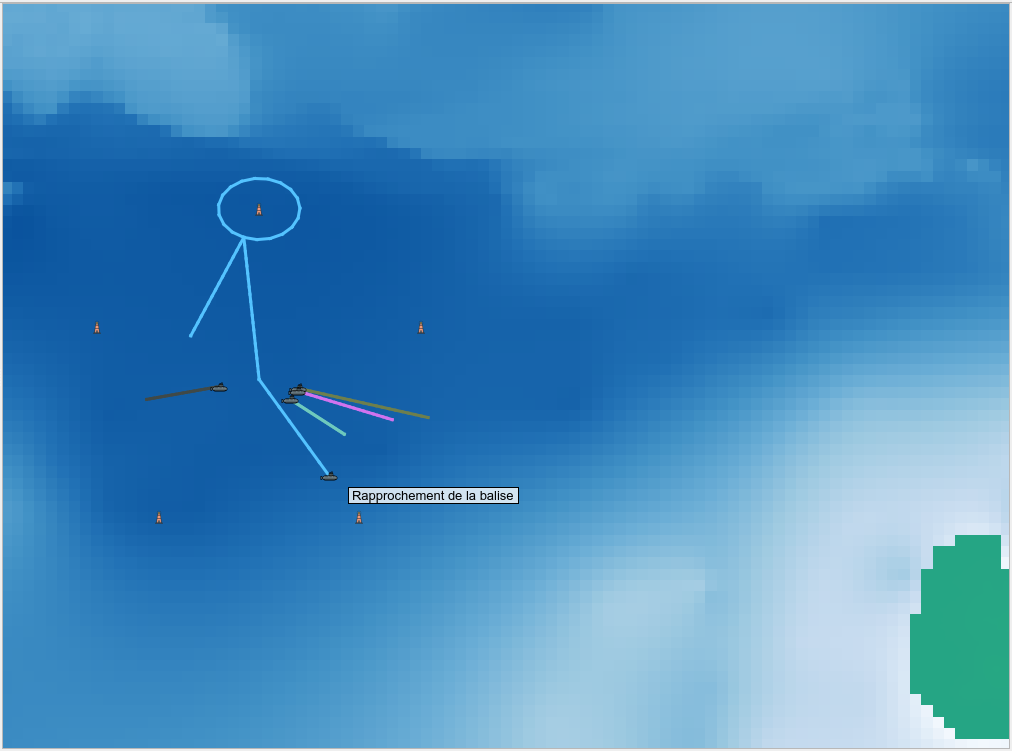}%
\label{subfig:d}%
}
\caption{A mission example executed in TraceQGIS illustrating the features of the navigation plugin.}
\label{fig:fig}
\end{figure*}

 A \emph{planning problem} $\cal{P}$ is a tuple $({\cal{D}}, {s_0}, w_0, g)$, where $\cal{D}$ is a planning domain, $s_0 \in S$ is the initial state, $w_0$ is the initial task network, and $g$ is a formula (not necessary ground) describing the goal.

 The HTN planning approach is particularly effective in scenarios where tasks must be dynamically decomposed into differing sets of subtasks, depending on the current state and problem constraints.
 Temporal dynamics, however, are not natively modelled.
Being able to represent hierarchical domains with time constraints is central for the execution of collaborative robotic missions in complex environments as it includes in the solution provided the necessary synchronization and coordination time points for the fleet.
Currently, there is no language for hierarchical planning with temporal constraints that is widely accepted as a standard within the community. Nevertheless, we propose here to adopt an early version of HDDL2.1 \cite{pellier2023hddl} -- a 2023 syntax and semantic framework for describing Temporal HTN problems -- which we intend to evolve further to meet our expressivity requirements.
As a first step, we express the planning problem in an untimed HDDL 1.0 model \cite{holler20}, developing it further to include time constraints in HDDL 2.1.
It is worth noting that few planners operate under this paradigm, and the development of such planner would in itself advance the state of the art.

\section{Implementation}
\label{sec:Implem}
All the elements of the software toolchain
have been implemented as QGIS extensions; QGIS (Quantum Geographic Information System) is a free, open-source software that allows users to create, edit, visualize, analyse, and publish geospatial information.~\cite{QGISsoftware}.

The algorithm implementing the deployment solution for the beacons helping the underwater localization of the fleet of UUVs can be easily implemented with A* search.

For the HTN planner, an off-the-shelf planner can be used to provide a hierarchical plan for the fleet, e.g. SHOP2~\cite{nau03}, or the more recent PANDA~\cite{panda}. A Temporal HTN planner to include temporal constraints in the navigation plan is currently under development.

The plugin TraceQGIS permits to load model files describing the HTN planning problem (in the HDDL language~\cite{pellier2023hddl,holler20}), and a configuration file (in YAML format) describing the visual interface, e.g. the icons used.
All entities participating in the mission (the beacons and the UUVs), their trajectories, and the ongoing communication are visualised in the QGIS interface in separate layers, namely \textit{ Entity, Communication, Trace Movements}  (see Figure~\ref{fig:interface}).

A simple scenario is used here to illustrate the capabilities of TraceQGIS with our approach for GNSS-denied underwater navigation. The mission consists in having a UUV, which initial position is uncertain, to navigate to beacon 8 and to broadcast a message half way between the 5 deployed beacons. The initial situation is illustrated in Figure~\ref{fig:interface}: the 5 beacons are deployed in the mission area, while the 5 UUVs are ``between the beacons'', with some uncertainty about their position. Beacon no. 6 is emitting an acoustic signal.

UUV no. 1 navigates towards the supposed position of beacon no. 6 (Fig.~\ref{subfig:a}), until it receives its acoustic signature (Fig.~\ref{subfig:b}). It then circles the beacon to localise itself precisely, and then broadcasts the message to the fleet (Fig.~\ref{subfig:c}). The message is received by the other UUVs, which then navigate towards the broadcasting position, while UUV no. 1 continues following its path to beacon no. 8 (Fig.~\ref{subfig:d}).

In the figures, the bathymetry data are downloaded from the Service hydrographique et océanographique de la marine (SHOM)~\cite{SHOM}.\vfill

\section{Conclusions and future work}
\label{sec:conclu}

We have presented an alternative solution for enabling stealthy underwater navigation operations using a fleet of autonomous UUVs. The proposed approach involves first generating an optimal deployment plan for marine acoustic beacons, accounting for challenges deriving from submarine operations. The beacons signal a navigation path, which the UUVs then follow via a sequence of actions synthesized by a hierarchical planner.
Continuous monitoring the plan execution
allows to adaptive replan in response to environmental perturbations, in order to dynamically adjust the plan based on real-time fleet state data, including positional estimates, beacon pose uncertainty, and the signals gathered during the navigation, reinforcing the robustness of the proposed solution.

%
Some aspects of the approach are left for future work.
The fleet coordination has not been considered in the first navigation model, assuming a parallel and independent navigation for the UUVs.
In the future, the inherent limits of submarine communication affecting the beacons signals and the inter-UUVs communications must be taken into account for a better solution.
%
%
In fact, the unknown underwater target position problem can be solved using trilateration~\cite{bayat2015range}, using UUVs following different trajectories and inter-UUV communication to compute in real time an estimate of the  ranges between the UUVs and simultaneously localize the beacons.

\vfill\null

\bibliographystyle{ieeetr}
\bibliography{biblio}

\begin{thebibliography}{00}
\bibitem{b1} G. Eason, B. Noble, and I. N. Sneddon, ``On certain integrals of Lipschitz-Hankel type involving products of Bessel functions,'' Phil. Trans. Roy. Soc. London, vol. A247, pp. 529--551, April 1955.
\bibitem{b2} J. Clerk Maxwell, A Treatise on Electricity and Magnetism, 3rd ed., vol. 2. Oxford: Clarendon, 1892, pp.68--73.
\bibitem{b3} I. S. Jacobs and C. P. Bean, ``Fine particles, thin films and exchange anisotropy,'' in Magnetism, vol. III, G. T. Rado and H. Suhl, Eds. New York: Academic, 1963, pp. 271--350.
\bibitem{b4} K. Elissa, ``Title of paper if known,'' unpublished.
\bibitem{b5} R. Nicole, ``Title of paper with only first word capitalized,'' J. Name Stand. Abbrev., in press.
\bibitem{b6} Y. Yorozu, M. Hirano, K. Oka, and Y. Tagawa, ``Electron spectroscopy studies on magneto-optical media and plastic substrate interface,'' IEEE Transl. J. Magn. Japan, vol. 2, pp. 740--741, August 1987 [Digests 9th Annual Conf. Magnetics Japan, p. 301, 1982].
\bibitem{b7} M. Young, The Technical Writer's Handbook. Mill Valley, CA: University Science, 1989.
\end{thebibliography}

\end{document}